\documentclass[letterpaper]{article} % DO NOT CHANGE THIS
\usepackage{aaai24}  % DO NOT CHANGE THIS
\usepackage{times}  % DO NOT CHANGE THIS
\usepackage{helvet}  % DO NOT CHANGE THIS
\usepackage{courier}  % DO NOT CHANGE THIS
\usepackage[hyphens]{url}  % DO NOT CHANGE THIS
\usepackage{graphicx} % DO NOT CHANGE THIS
\usepackage{natbib}  % DO NOT CHANGE THIS AND DO NOT ADD ANY OPTIONS TO IT
\usepackage{caption} % DO NOT CHANGE THIS AND DO NOT ADD ANY OPTIONS TO IT
\usepackage{algorithm}
\usepackage{algorithmic}

\usepackage{algorithm}
\usepackage{xcolor}
\usepackage{graphicx}
\usepackage{color,soul}
\usepackage{amsmath}
\usepackage{amssymb}

\usepackage{newfloat}
\usepackage{listings}
\DeclareCaptionStyle{ruled}{labelfont=normalfont,labelsep=colon,strut=off} % DO NOT CHANGE THIS
\floatstyle{ruled}
\newfloat{listing}{tb}{lst}{}
\floatname{listing}{Listing}

\nocopyright 
\usepackage{bibentry}

\begin{document}

\title{\texttt{SALSA}: Semantically-Aware Latent Space Autoencoder}
\author{
    Kathryn E. Kirchoff,\textsuperscript{\rm 1}, 
    Travis Maxfield,\textsuperscript{\rm 1}, 
    Alexander Tropsha,\textsuperscript{\rm 1}\thanks{Corresponding author},
    Shawn M. Gomez\textsuperscript{\rm 1}\thanks{Corresponding author}
}
\affiliations{
    \textsuperscript{\rm 1}University of North Carolina\\
        kat@cs.unc.edu, 
        tmaxfield@unc.edu, 
        alex\_tropsha@unc.edu, 
        smgomez@gmail.com
}

\maketitle
\begin{abstract}

In deep learning for drug discovery, chemical data are often represented as simplified molecular-input line-entry system (SMILES) sequences which allow for straightforward implementation of natural language processing methodologies, one being the sequence-to-sequence autoencoder. However, we observe that training an autoencoder solely on SMILES is insufficient to learn molecular representations that are semantically meaningful, where semantics are defined by the structural (graph-to-graph) similarities between molecules. We demonstrate by example that autoencoders may map structurally similar molecules to distant codes, resulting in an incoherent latent space that does not respect the structural similarities between molecules. To address this shortcoming we propose \underline{S}emantically-\underline{A}ware \underline{L}atent \underline{S}pace \underline{A}utoencoder (\texttt{SALSA}), a transformer-autoencoder modified with a contrastive task, tailored specifically to learn graph-to-graph similarity between molecules. Formally, the contrastive objective is to map structurally similar molecules (separated by a single graph edit) to nearby codes in the latent space. To accomplish this, we generate a novel dataset comprised of sets of structurally similar molecules and opt for a supervised contrastive loss that is able to incorporate full sets of positive samples. We compare \texttt{SALSA} to its ablated counterparts, and show empirically that the composed training objective (reconstruction and contrastive task) leads to a higher quality latent space that is more 1) structurally-aware, 2) semantically continuous, and 3) property-aware.

\end{abstract}

% % % % % % % % % % % % % % % % % % % % % % % % % % % % % %

\section{Introduction} \label{sec:intro}

    \begin{figure}[t!]
        \centering
        \includegraphics[width=1\columnwidth]{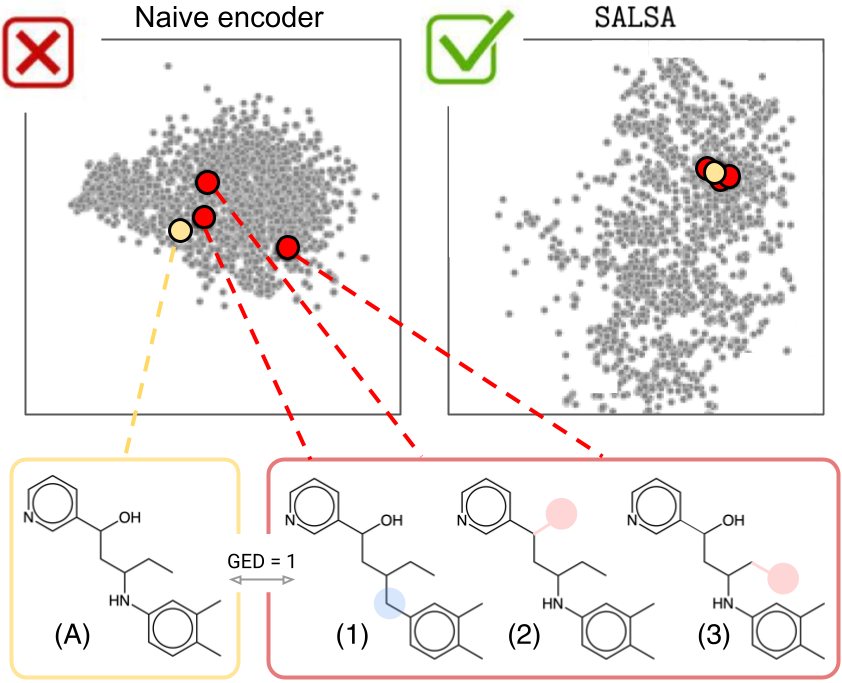}
        \caption{Given molecule (A) we consider three molecules whose graphs are structurally similar, being a single graph edit from (A). The naive autoencoder maps these similar molecules to latent codes of various proximity: (1) is mapped close to (A), while (3) is mapped far from (A). In contrast, our proposed autoencoder, \texttt{SALSA}, learns a semantically organized space such that structurally similar molecules are collectively mapped to nearby codes.}
        \label{fig:FigOfInterest}
    \end{figure}%[h]

    In drug discovery, learning the underlying semantics that govern molecular data presents an interesting challenge for deep learning. Effective learning of semantics is necessary to be successful in key tasks such as property prediction and \textit{de novo} generation, and progress has been made in attempting to solve these tasks \cite{gen_model_rev}. However, due to the ambiguous nature of molecular representations, models often fail to adequately capture the underlying semantics resulting in a disorganized latent space.

    In the case of molecular data, semantics is often task-dependent but may amount to various emergent chemical attributes intrinsically linked to molecular structure, or the arrangement of constituent atoms and bonds \cite{qsar}. As molecular structure can be captured in the form of a graph, the semantics that govern chemical manifolds may therefore be defined by the graph-to-graph similarities (i.e. structural similarities) between molecules; we further claim that graph edit distance (GED) defines a semantically meaningful unit of change between molecular entities. 

    To take advantage of recent progress in sequence-to-sequence modeling, molecules may be expressed as text sequences, known as simplified molecular-input line-entry system (SMILES) strings \cite{Weininger1988}. Borrowing from advancements made in natural language processing (NLP), many autoencoder-based methods that use SMILES sequences have been proposed as they provide a promising framework to solve problems in drug discovery~\cite{alperstein2019smiles, chemvae, gen_model_rev}.

    However, these models are plagued by some of the same challenges met in the field of NLP, namely, difficulties in learning latent spaces that capture underlying sentence semantics \cite{simcse, shen2020educating}. 
    This arises from the fact that for discrete objects such as sentences, autoencoders have the capacity to map similar data to distance latent representations. 
    We observe an analogous problem in that SMILES-based autoencoders are not able to adequately learn the structure-based semantics that underlie chemical datasets, and as a result, these models may map semantically similar molecules to distant latent codes in the latent space. This phenomenon is more precisely defined as an instance in which semantically similar molecules (low GED) have are mapped to distant latent representations (high Euclidean distance). We show an example of this in Figure \ref{fig:FigOfInterest}. Collectively, many of these semantically naive events induce a disorganized latent space which limits success of these models in downstream tasks.

    To remedy this shortcoming of SMILES-based autoencoders, we propose enforcing a sense of semantic awareness on to an autoencoder such that structurally similar molecules are mapped near one another in the latent space. Our proposed model, \underline{S}emantically-\underline{A}ware \underline{L}atent \underline{S}pace \underline{A}utoencoder (\texttt{SALSA}), is a modified SMILES-based transformer autoencoder that, in addition to a canonical reconstruction loss, learns a contrastive task having the objective of mapping structurally similar molecules, whose graphs are separated by a single edit distance, to similar codes in the effected latent space. In this way, we are able to learn a semantically meaningful and continuous latent space. We compare \texttt{SALSA} to its two ablations (a naive SMILES autoencoder and a contrastive encoder) and evaluate their latent spaces in terms of structural awareness, semantic continuity, and property awareness. We are the first, to our knowledge, to enforce structural (graph-based) awareness onto a SMILES-based model.

    \subsubsection{Our contributions are as follows:}
    \begin{itemize}
        \item We propose a novel modeling framework, \texttt{SALSA}, that composes a transformer-based autoencoder with a contrastive task to achieve semantically-aware molecular representations.
        \item We develop a scheme for constructing a chemical dataset suited to contrastive learning of molecular entities, specifically aimed at learning graph-to-graph similarities between molecules. 
        \item We evaluate the quality of \texttt{SALSA}'s latent space based on: 1) structural awareness, 2) semantic continuity, and 3) physicochemical property awareness.  
    \end{itemize}

\section{Related Works} \label{sec:background} 

    \subsubsection{Sequence-based Models}
        Among the various modes of molecular representation, a sequence-based, i.e.\ SMILES-based, representation is uniquely able to benefit from advances in sequence-to-sequence modeling, especially those following the introduction of the transformer architecture~\cite{attention_is_all_you_need}. 
        While the original transformer architecture comprised an encoder-decoder pair, it did not provide a global representation of the input. Earlier work aimed at embedding whole sequences into a latent space used recurrent neural networks, which are more naturally aligned to this objective \cite{Bowman_Vilnis_Vinyals_Dai_Jozefowicz_Bengio_2016, shen2020educating}. Since then, other authors have modified the transformer architecture to include a bottleneck (or pooling) layer and a single, fixed-size global embedding of the input \cite{montero-etal-2021-sentence,9054554,li-etal-2020-optimus}. Several authors have applied sequenced-based models to molecular representations. Examples of RNN-based models include ChemVAE \cite{chemvae} and the All SMILES VAE \cite{alperstein2019smiles}. Transformer-based models include ChemBERTa \cite{Chithrananda2020-wf} and SMILES Transformer \cite{honda2019smiles}.

    \subsubsection{Contrastive Learning}        
        For molecular data, both SMILES and graph representations have been explored in the context of contrastive learning. The model proposed by \cite{FragNet} utilized the normalized temperature-scaled cross entropy (NT-Xent) \cite{ntxent} loss to map enumerated SMILES of identical molecules nearby in the latent space. Insofar as graphs, \citeauthor{Wang2021-we} similarly used the NT-Xent loss to maximize the agreement between pairs of augmented graphs (``views'') describing the same molecule; here, each view (i.e. positive sample) is obtained by masking out nodes or edges. The NT-Xent loss, although widely successful, operates solely on positive \textit{pairs}, an issue addressed by \cite{SupCon} in their formulation of the Supervised Contrastive (SupCon) loss which allows for comparison among an arbitrarily sized \textit{set} (rather than a pair) of positive instances.

\section{Methodology} \label{sec:methods}
    \subsection{Overview of Approach}
        Broadly, our goal is to modify an autoencoder, operating on SMILES inputs, such that the effected latent representation better respects the structural similarities, particularly the graph edit distance (GED), between molecular pairs. We do this by incorporating a contrastive component into the architecture.

        \subsubsection{Contrastive Component}
            The contrastive task necessitates labeled pairings of ``similar'' and ``dissimilar'' molecules. We opt to consider as similar any two molecules separated by a single graph edit. Recall that the graph edit distance (GED) between two molecules, viewed as labeled graphs, is the minimum number of single edits required to make one graph isomorphic to the other. It is computationally infeasible to obtain all pairs of single GED molecules systematically from an existing dataset. To sidestep this issue, we generate a bespoke dataset of 1-GED molecular pairings. We accomplish this by defining a set of single graph-edit transformations, or mutations, which are applied to ``anchor'' molecules to obtain similar molecules which we will refer to as ``mutants''.
        \subsubsection{Autoencoder Component}
            We define our autoencoder with a transformer-based encoder and decoder, and an intermediate bottleneck to produce a latent embedding space. Combined with the contrastive component, the general framework is encapsulated in figure \ref{fig:Architecture}. We note that an encoder trained solely on the contrastive objective, that is, without the reconstruction loss central to an autoencoder, may learn a degenerative mapping such that our designated ``similar'' molecules are mapped to representations that are in fact \textit{too similar}, being almost stacked on top of one another. In this way, the reconstruction loss provided by the autoencoder component acts as a regularizer to encourage similar molecules to be mapped to distinct codes.

    \subsection{Training Dataset} \label{sec:training_dataset} 

        \subsubsection{Anchor Compounds}
            We utilize the dataset developed by \cite{Popova2018}, which contains approximately 1.5 million SMILES sequences sourced from the ChEMBL database (version ChEMBL21), a chemical database comprised of drug-like or otherwise biologically-relevant molecular compounds \cite{ChEMBL2014}. 
            After procuring the full dataset, the set of compounds was run through a standard curation pipeline; for an in-depth description of the curation process, please refer to \cite{Popova2018}. 
            We further filter the dataset by only allowing molecules whose SMILES sequence length was less than or equal to 110 characters, leaving 1,256,277 compounds. These compounds constitute the anchors from which we generate 1-GED mutants, as further explained in the following subsection.

        \begin{figure}[t!]  %[h]
            \centering
            \includegraphics[width=.95\columnwidth]{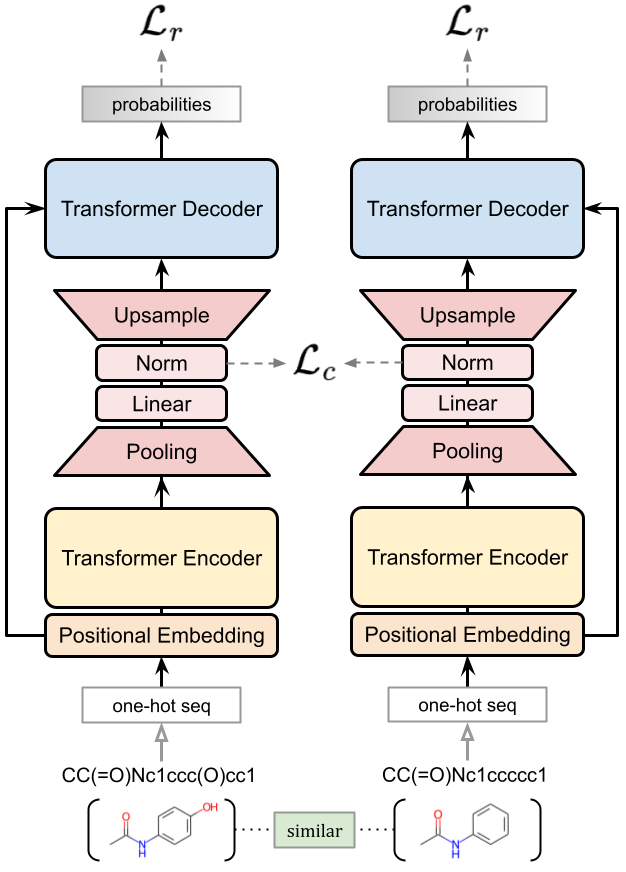}
            \caption{Overview of \texttt{\texttt{SALSA}} architecture. \texttt{\texttt{SALSA}} operates on multi-mutant batches, but here, we show a single (positive) anchor-mutant pair for simplicity. The reconstruction loss ($\mathcal{L}_r$) is computed on SMILES representations. In the case of a positive pair (similar molecules) as shown, the contrastive loss ($\mathcal{L}_c$) aims to map their normalized representations to nearby codes in the latent space. Note that weights between the two networks are shared, thus, only a single model is trained and used for inference.}
            \label{fig:Architecture}
        \end{figure}%[h]

        \subsubsection{Mutant Compounds}\label{sec:mutant_generation} 
            
            We define a molecular graph generally as $g = (\mathcal{V}, \mathcal{E})$ where $\mathcal{V} = \{v_0, ..., v_A\}$ is the set of nodes, where each $v_a \in \{\texttt{C}, \texttt{O}, \texttt{N}, \texttt{S}, \texttt{Br}, \texttt{Cl}, \texttt{I}, \texttt{F}, \texttt{P}, \texttt{B}, \texttt{\$}\} $ (atom types), and $\mathcal{E} = \{(v_a, v_b)| v_a, v_b \in \mathcal{V}\}$ is the set of edges (bonds). Note that atom type $\texttt{\$}$ is a stand-in for any atom type not in the remaining list, analogous to an $\texttt{unk}$ character in natural language models. 
            
            Here, we will differentiate anchors from mutants with a tilde, i.e. anchor graphs as $g$ and mutant graphs as $\tilde{g}$. 
            Given an anchor, we consider its graph, $g_{i} \in G$ where $G$ is the anchor set (sourced from ChEMBL) and $i$ is the index identifying the anchor in $G$. We obtain a mutated graph, or mutant, by randomly sampling a mutation operator $t(\cdot) \sim \mathcal{T}$ and applying that mutation to the anchor, $t(g_{i}) = \tilde{g}_{i(j)}$ where $i$ again corresponds to the original anchor, and $j$ is the index of the mutant graph within the anchors' positive sample set. 
            
            The set of mutation operators, $\mathcal{T}$, is defined to avoid mutations that would drastically alter the graph structure, i.e.\ separating molecules into disconnected graphs or breaking and forming rings. Furthermore, we require mutants to be chemically valid molecular graphs, and we normalize all SMILES using the RDKit canonicalization algorithm~\cite{rdkit}. In defining these mutations, we keep hydrogen atoms implicit, adding or removing as needed to satisfy chemical validity. Our mutation operators are defined as follows. 
            \begin{itemize}
                \item \textit{Node addition} (\texttt{add}): Append a new node, and a corresponding edge, to an existing node in the graph.
                \item \textit{Node substitution} (\texttt{replace}): Change the atom type of an existing node in the graph. 
                \item \textit{Node deletion} (\texttt{remove}): Remove a singly-attached node and its corresponding edge from the graph.  
            \end{itemize}
            For both \texttt{add} and \texttt{replace}, incoming atom types are drawn from the observed atom type distribution in the original ChEMBL dataset. 
            % shown in Table \ref{tab:atom-counts}. 
            We have now defined our curated set of node-level graph transformations, $\mathcal{T} = \{\texttt{remove}, \texttt{replace}, \texttt{add}\}$. For each anchor, $g_i$, we generate 10 distinct mutants that constitute the ``positive'' sample set, $P(i)$, for that anchor:
            \begin{equation}
                P(i) = \{\tilde{g}_{i(1)}, \tilde{g}_{i(2)}, \dots, \tilde{g}_{i(10)}\}  \in \tilde{G}
            \end{equation}
            
            The resulting training set was composed of the original anchor compounds and their respective mutant compounds, amounting to 13,819,047 total training compounds.

            \begin{figure}[t!]
                \centering
                \includegraphics[width=1.\columnwidth]{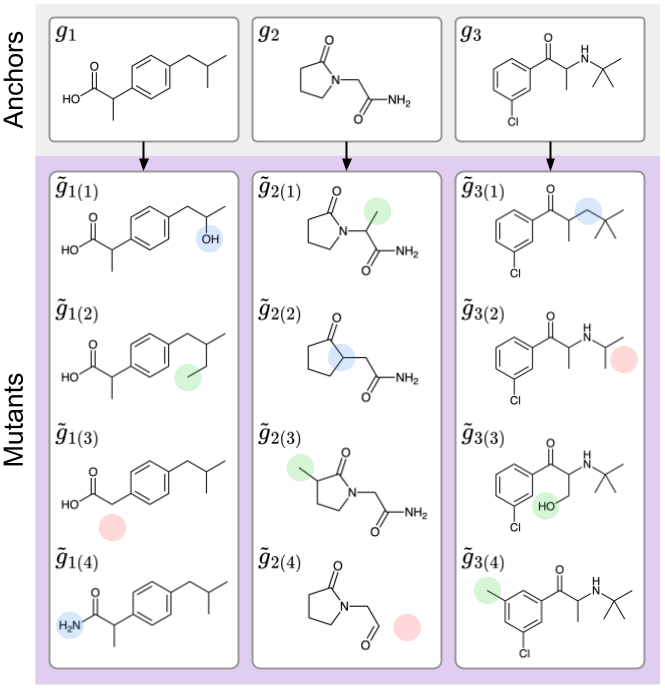}
                \caption{Multi-mutant batching of three anchors $g_i$ where $i \in \{1,2,3\}$ and for each anchor, a set of positive samples (1-GED mutants) denoted $P(i) = \{\tilde{g}_{i(n)}, \tilde{g}_{i(2)}, \tilde{g}_{i(3)}, \tilde{g}_{i(4)}\}$. For each anchor, the negatives samples are those mutants belonging to either of the other two anchors. Colored atoms correspond to single graph edits from anchor to mutant: \texttt{add} (green), \texttt{replace} (blue), and \texttt{remove} (red).}
                \label{fig:Mutants}
            \end{figure}%[h]

        \subsubsection{Faulty-Positive Filtering}
            Although our mutation operators ensure chemical validity, they do not ensure physicochemical proximity of mutants to anchors. Due to the complex nature of quantum mechanics underlying molecular interactions, a single graph edit mutation may effect great differences in the physicochemical properties between anchor and mutant. To mitigate such cases, we remove mutants that are too dissimilar from their respective anchor according to a physicochemical property threshold. In this case, dissimilarity is defined as the Mahalanobis distance between an anchor's and a mutant's respective sets of physicochemical properties. The Mahalanobis distance, $d_M( \cdot ,\cdot)$, between an anchor $g_i$ and its mutant $\tilde{g}_{i(j)}$ is computed as
            \begin{equation}
                d_M(g_i,\tilde{g}_{i(j)}) = \sqrt{(x_i - \tilde{x}_{i(j)})^{\intercal} \Sigma^{-1} (x_i - \tilde{x}_{i(j)})}
            \end{equation}
            where $x_i$ is the vector of physiochemical properties calculated for $g_i$, and  $\tilde{x}_{i(j)}$ is the analogous property vector for $\tilde{g}_{i(j)}$, and $\Sigma$ is the covariance matrix of the distribution of physicochemical properties over our collection of anchors, drawn from ChEMBL. 

    \subsection{Modeling Framework} \label{sec:modeling_framework}
    
        \subsubsection{\texttt{SALSA} Architecture} 
            The core architecture of \texttt{SALSA} is based on the transformer paradigm proposed by \citeauthor{NIPS2017_3f5ee243}, an encoder paired with an autoregressive decoder. \texttt{SALSA} operates on SMILES sequences corresponding to chemical graphs. A SMILES string is an ordered list of the atom and bond types encountered during a depth-first traversal of a spanning tree of the associated molecular graph (e.g. the SMILES sequence of ibuprofen is ``\texttt{CC(Cc1ccc(cc1)C(C(=O)O)C)C}''). 
            % We adopted a simple tokenization strategy yielding a vocabulary of 39 tokens, including the most common atom and bond types present in drug-like organic molecules in addition to the start token ,``$<$'', the end token, ``$>$'', the pad token, ``X'', and the unknown token, ``\$'', which is used in cases where \texttt{SALSA} encounters an atom type not present in the provided vocabulary.
            We adopted a simple tokenization strategy yielding a vocabulary of 39 tokens, including the most common atom and bond types present in drug-like organic molecules in addition to a start token, end token, pad token, and an unknown token, used in cases where \texttt{SALSA} inputs contain a token not present in the provided vocabulary.
            
            We modify the original transformer architecture by introducing a pooling layer and a subsequent upsampling layer between the encoder and decoder, and in this way impose an autoencoder framework. 
            The intermediate pooling layer limits the decoder's input such that the decoder receives a fixed-sized latent representation, rather than all contextualized embeddings produced by the encoder.
            Specifically, whereas the intermediate output of the transformer encoder is a vector of size $\mathbb{R}^{L\times H}$ for a sequence of length $L$ and hidden dimension size $H$, \texttt{SALSA} is designed to output a latent vector of fixed size $\mathbb{R}^S$. This is accomplished by first applying a component-wise mean pooling from $\mathbb{R}^{L\times H} \to \mathbb{R}^{H}$ before projecting $\mathbb{R}^{H} \to \mathbb{R}^{S}$.
            
            The \texttt{SALSA} latent vector is constrained to live on the unit hypersphere embedded in $\mathbb{R}^{S}$, and so we therefore normalize the output of the ``Pooling'' layer before subsequent manipulations. These manipulations include routing to the contrastive loss function, to be explained further below, and into the transformer decoder. As the transformer decoder is designed to accept an input of size equal to the output of the encoder, i.e. $\mathbb{R}^{L\times H}$, we first pass the latent vector through a linear layer with the appropriate output dimension and reshape as needed. This is referred to as ``Upsample'' in Figure \ref{fig:Architecture}. Note that this method of injecting the latent vector into the transformer decoder resembles the method called `memory' in~\cite{li-etal-2020-optimus}, where it was demonstrated to yield superior results over an alternative strategy.

        \subsubsection{Loss Function} \label{loss_function}
            We define a compound loss function, composed of: (1) a contrastive component defined over a batch of inputs, and (2) a canonical reconstruction component, characteristic of autoencoder neural networks. For the contrastive component, we adapt the supervised contrastive (SupCon) loss \cite{SupCon}. The SupCon loss allows for multiples positive comparisons per anchor, resulting in improved performance relative to naive contrastive losses, which operate on the assumption of only a single positive sample per anchor. The SupCon is defined as 
            \begin{equation}
                \mathcal{L}_c = \sum_{i \in I}  \frac{-1}{|P(i)|} \sum_{p \in P(i)}\log \frac{\exp{(z_i \cdot z_p}/ \tau)}{\sum_{a \in A(i)} \exp{(z_i \cdot z_a}/ \tau)},
            \end{equation}
            where $A(i)$ is the set of all samples sharing a batch with instance $i$, with latent code $z_i$, and $P(i)$ are those elements of $A(i)$ that are similar to $i$, and $I$ is the set of anchors in the batch, using the terminology of Sec. \ref{sec:training_dataset}. 

            For our reconstruction loss, we opt for causal masking, giving the following formulation in terms of cross-entropy for a single sequence $s_i$ and its associated latent vector $z_i$: 
            \begin{equation}
                \mathcal{L}_{r,i}(s_i) = -\frac{1}{T}\sum_{t=1}^{T} \log p_\theta(s_{i,t}|z_i, s_{<t,i}),
            \end{equation}
            where we've let $T$ be the length of the sequence $s_i$ and denoted as $ p_\theta(s_{i,t}|z_i, s_{<t,i})$ the output of the decoder at position $t$ along the sequence. Note the dependence on $z_i$, the latent code---output of the encoder---associated to the sequence $s_i$. The full reconstruction loss $\mathcal{L}_r$ is the average of all per-sequence losses.
            
            The final loss computation is simply a weighted combination of the two defined above.
            \begin{equation}
                  \mathcal{L} = \lambda \mathcal{L}_c + (1- \lambda) \mathcal{L}_r
            \end{equation}
            where $0\leq \lambda \leq 1$ is a hyperparameter for weighting the contributions of the contrastive loss and the reconstruction loss, respectively. We train \texttt{SALSA} with $\lambda=0.5$, and make comparisons to either ablation,\ $\lambda = 1$ and $\lambda = 0$ described in Sec. \ref{sec:experiments}. 

        \subsubsection{Implementation Details}
             We use $l=8$ layers for both the encoder and the decoder with a hidden dimension of size $h = 512$, and $m = 8$ heads in the multi-head attention blocks. Our main results are of models trained with $S = 32$ latent dimensions, although we also investigated reduced latent dimensions, $S \in \{16,8,4,2\}$. For the contrastive loss, we set temperature $\tau = 0.7$, which is the default.

\section{Experiments and Analysis} \label{sec:experiments}
            
    We are interested in gathering a comprehensive understanding of \texttt{SALSA}'s latent space relative to both a naive autoencoder and a contrastive encoder. To this end, we ask three questions of our \texttt{SALSA} representations: 
        
    \begingroup
        \renewcommand\labelenumi{(\theenumi)}
        \begin{enumerate}
            \item \textbf{Structural Awareness:} Does \texttt{SALSA} encode information about structural (graph-to-graph) relationships? 
            \item \textbf{Semantic Continuity:} Does \texttt{SALSA} produce interpolations that are more semantically reasonable?
            \item \textbf{Property Awareness:} Does \texttt{SALSA} implicitly encode information about physicochemical properties? 
        \end{enumerate}
    \endgroup
    \noindent

    \subsection{Baselines}
        \begin{itemize}
            \item \textbf{Naive Autoencoder:} We are interested in \texttt{SALSA}'s performance relative to a naive autoencoder trained solely on SMILES reconstruction. To obtain a naive autoencoder, we trained \texttt{SALSA} with weighting hyperparameter, $\lambda = 0$. We abbreviate this model as ``Naive''.
            \item \textbf{Contrastive Encoder:} We are also interested in how the reconstruction objective influences the effectiveness of the contrastive task in achieving structural awareness. To obtain a contrastive encoder, we trained \texttt{SALSA} with weighting hyperparameter, $\lambda = 1$. We abbreviate this model as ``Contra''.
        \end{itemize}

    \subsection{Structural Awareness}

    In order to evaluate the degree to which representations capture structural awareness, we compute metrics of correlation between Euclidean distance (EuD) in the latent space and graph edit distance (GED) in the data space. Correlation metrics necessitate \textit{a priori} knowledge of GEDs between molecular pairs of interest, not unlike the anchor–mutant pairs generated for our training set. Thus, we extend our mutation process to generate sets of mutants having known GEDs (one to five) from their anchors, which we will refer as ``supermutants''.

        \begin{figure}[t!]
            \centering
            \includegraphics[width=1.\columnwidth]{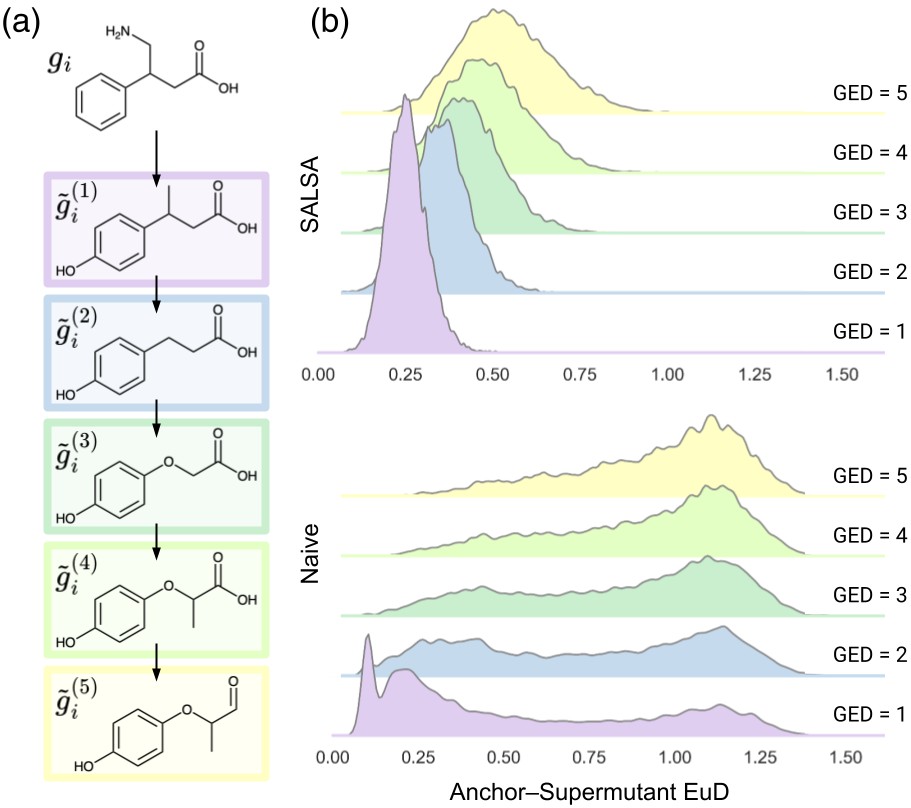}
            \caption{
            % Supermutants and pairwise distances in latent space. Supermutants are color-coded according to $n$-GED (1-GED: purple, 2-GED: blue, 3-GED: green, etc.) from the anchor; color-coding is consistent across both sub-figures. 
            (a) Example of supermutants, $S(i)$, generated from an anchor, $g_i$. (b) Anchor–supermutant Euclidean distances, by $n$-GED, in Naive and \texttt{SALSA} latent spaces. For both subfigures, supermutants are color-coded according to $n$-GED (1-GED: purple, 2-GED: blue, 3-GED: green, etc).}
            \label{fig:Supermutants}
        \end{figure}%[h]

        \begin{figure*}[t!]
            \centering
            \includegraphics[width=1.\textwidth]{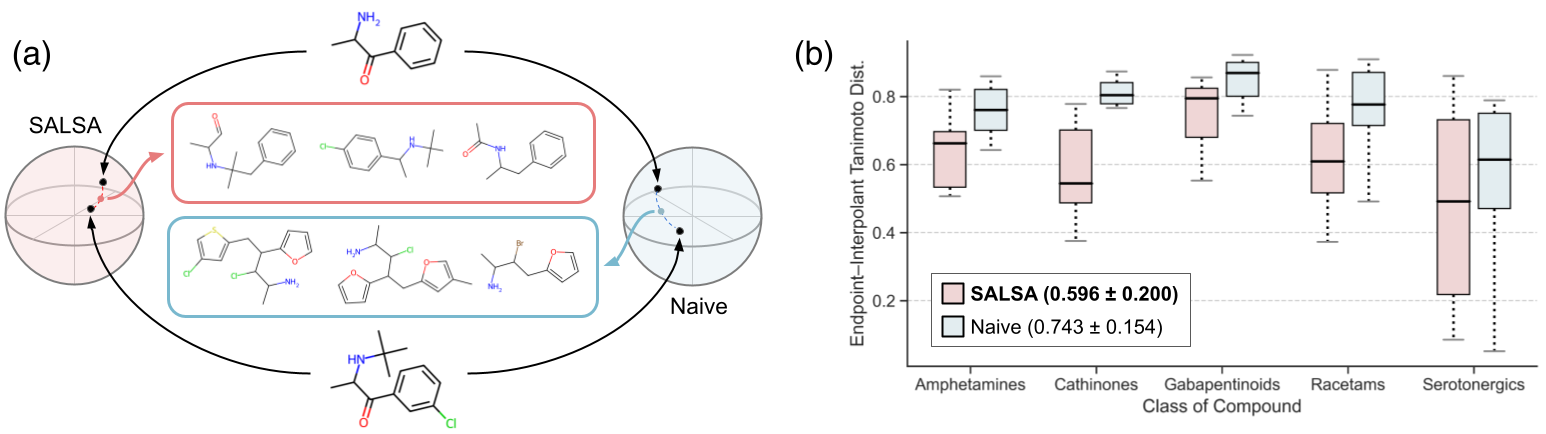}
            \caption{(a) Three most common midpoint interpolants between cathinone (top) and bupropion (bottom), generated from either \texttt{SALSA} space or Naive space. (b) Box plots showing the distribution of endpoint-interpolant Tanimoto distances per compound class for both Naive and \texttt{SALSA} (lower is better). Legend shows overall mean and standard error.}
            \label{fig:Interps}
        \end{figure*}%[h]

        \subsubsection{Supermutant Compounds}
            We already demonstrated our ability to generate 1-GED mutants, but here we generate sets of $n$-GED supermutants where $n \in \{1,2,3,4,5\}$. 
            % To this end, we made use of our set of mutation operators, $\mathcal{T}$, 
            For a given anchor, $g_i$, we apply a random mutation, $ t^{(1)}(\cdot) \sim \mathcal{T}$, to generate a 1-GED (super)mutant, $\tilde{g}_i^{(1)} = t^{(1)}(g_i)$, to which another random mutation operator is applied to generate a 2-GED supermutant, $\tilde{g}_i^{(2)} = t^{(2)}(\tilde{g}_i^{(1)}) $, and so-on. Applying $n$ randomly chosen mutations yields an $n$-GED supermutant with high probability. Through this process, we generate sets of molecules having known $n$-GEDs, $n \in \{1,2,3,4,5\}$, from each respective anchor molecule.\footnote{Note that superscripts here, e.g.\ $\tilde{g}^{(2)}_i$, are distinct from the subscripts labeling mutants in Sec. \ref{sec:training_dataset}, e.g.\ $\tilde{g}_{i(2)}$. Here the superscript denotes the number of mutations \emph{successively} applied to an anchor, while the subscript in Sec. \ref{sec:training_dataset} indexes the number of $1$-GED mutants used for training.}
            
            To be specific, for each anchor, we choose a single $1$-GED mutant which we mutate again to find a single $2$-GED mutant, and so on, so that each of the $n$-GED supermutants in our evaluation set is a mutation of an $(n-1)$-GED supermutant also in the evaluation set. We can express this as the following supermutant set depending on a chosen anchor $g_i$:
            
            \begin{equation}
                S(i) = \{\tilde{g}_i^{(1)}, \tilde{g}_i^{(2)},  \tilde{g}_i^{(3)},  \tilde{g}_i^{(4)}, \tilde{g}_i^{(5)}\}.
            \end{equation}
            
            These resulting sets of supermutants and their associated anchors comprise the evaluation dataset used to assess correlations between Euclidean distance in latent space and GED (between anchors and respective supermutants). Figure \ref{fig:Supermutants}(a) shows an example of successive supermutant molecules generated from an anchor molecule.

            Furthermore, to ensure that we are evaluating the robustness of our model's learned transformations rather than the extent of their training memorization, we opt for an anchor set that is independent of the training set. To this end, we utilize the ChEMBL23 dataset, which are all chemical compounds in ChEMBL curated as of May 2023. We take the difference between the ChEMBL23 dataset and our training anchors (originally pulled from the ChEMBL21 dataset, being all compounds collected up until March 2016) to obtain an independent set of anchors (randomly sampled to 5000) from which to generate supermutants.

        \subsubsection{GED-EuD Correlation}
        \begin{table}[t!] %[htbp] 
            \centering
            \begin{tabular}{lcc}
            \hline
            \hline
            Method & Spearman's $\rho$ & Kendall's $\tau$ \\ 
            \hline
            \hline
            Naive-32 & 0.514 $\pm$ 0.53 & 0.467 $\pm$ 0.48 \\
            \textbf{Contra-32} & \textbf{0.888 $\pm$ 0.20} & \textbf{0.841 $\pm$ 0.23} \\
            \textbf{\texttt{SALSA}-32} & \textbf{0.878 $\pm$ 0.20} & \textbf{0.824 $\pm$ 0.23} \\
            \hline
            \texttt{SALSA}-16  & 0.849 $\pm$ 0.24& 0.789 $\pm$ 0.26 \\
            \texttt{SALSA}-8   & 0.807 $\pm$ 0.28& 0.741 $\pm$ 0.30  \\
            \texttt{SALSA}-4   & 0.587 $\pm$ 0.46& 0.518 $\pm$ 0.42 \\
            \texttt{SALSA}-2   & 0.351 $\pm$ 0.57& 0.300 $\pm$ 0.50   \\
            \hline
            \end{tabular}
            \caption{Spearman's $\rho$ and Kendall's $\tau$ for GED-EuD correlation in latent space of various models. We compare \texttt{SALSA}, Contra, and Naive trained at $d=32$. We further compare \texttt{SALSA} models trained at reduced dimensions. The highest performing methods are in boldface.}
        \label{tab:correlation_comparison}
        \end{table}

            With our set of independent anchors and associated supermutants, we can evaluate the correlation between GED, $d_{\texttt{GE}}$, between molecular graphs and Euclidean distance, $d_{\texttt{Eu}}$ or EuD, in the latent space. For a given anchor, $g_i$, and one of its supermutants $\tilde{g}_i^{(n)}$:
            \begin{equation}\label{eq:eud}
                d_{\texttt{Eu}}(g_i,\tilde{g}_{i}^{(n)}) = \lVert z_i -  \tilde{z}_{i}^{(n)} \rVert ^2_2
            \end{equation}   
            \begin{equation}
                d_{\texttt{GE}}(g_i,\tilde{g}_{i}^{(n)}) = n
            \end{equation}   
            where $z_i$ and $\tilde{z}_{i}^{(n)}$ are the latent representations of the anchor and the supermutant, respectively. Eq. (\ref{eq:eud}) gives us 5000 EuDs at each $n$-GED depth $n \in \{1,2,3,4,5\}$, and we show the resulting distributions in Figure \ref{fig:Supermutants}(b) for Naive and {\texttt{SALSA}}. We note that the Contra distribution is practically indistinguishable from {\texttt{SALSA}} so, for brevity, it is not included in Figure \ref{fig:Supermutants}(b).

            We then calculate two measures of rank correlation, Spearman correlation coefficient ($\rho$) and Kendall correlation coefficient ($\tau$), between GED and EuD for each anchor-supermutant set, $i \in [1,5000]$. Then we compute the average Spearman $\rho$ and Kendall $\tau$ across all 5000 correlations. We perform this analysis on \texttt{SALSA}, Naive, and Contra space at 32 dimensions, and we further investigated \texttt{SALSA} performance at lower dimensions, $d \in \{16,8,4,2\}$. Results for this evaluation are shown in Table \ref{tab:correlation_comparison}. 
            
            For this analysis, \texttt{SALSA} and Contra are on par having the highest correlations (among the 32-$d$ spaces), compared to Naive. Although, \texttt{SALSA} does perform slightly worse than Contra, presumably due to the regularizing nature of the reconstruction task.  We note that Naive has a wide standard error that is revealed in further detail in Figure \ref{fig:Supermutants}(b). The bimodal distribution of Naive at $n$-GED may be interpreted as single graph edits inducing changes to SMILES strings that are either mild (the left mode) or vast (the right mode). $\texttt{SALSA}$ comparatively produces distributions that are consistently unimodal, although the distribution flattens with increasing $n$-GED indicating that the correlation may not hold between anchors and mutants that are substantially different.
            Lastly, we find that with decreasing dimensionality, \texttt{SALSA}'s performance does not significantly degrade until $d=4$, suggesting avenues for potential exploration into applications that necessitate operation in exceptionally small dimensional spaces.

    \subsection{Semantic continuity (Interpolations)}

        We investigate \texttt{SALSA}'s ability to generate reasonable molecular interpolations between pairs of endpoint molecules, as higher quality interpolations suggest better semantic continuity in the latent space \cite{shen2020educating}. 
        To obtain interpolations, we choose pairs of ``endpoint'' molecules, calculate the spherical linear interpolation (\textit{slerp}) midpoint \cite{sampling_gen} between them, and then feed the midpoint code to a decoder to stochastically generate interpolant molecules. We do not perform this evaluation on Contra as it lacks a decoder for generating.
        Figure \ref{fig:Interps}(a) shows a case study interpolating between two molecules to get the three most common midpoint interpolants generated by both the \texttt{SALSA} decoder and the Naive decoder. Qualitatively, we see that \texttt{SALSA} generates interpolants that are much more structurally similar to the endpoints.
        
        Next, we quantify \texttt{SALSA}'s interpolation capability more comprehensively. To this end, we consider five classes of compounds (serotonergics, gabapentinoids, amphetamines, cathinones, and racetams), and for each class, choose sets of four to five molecules. For each set, we consider all pairwise combinations and for each pair, find the most common midpoint interpolant. 
        Then we calculate the Tanimoto distance, a common measure of chemical similarity used in medicinal chemistry, between each interpolant and either of their endpoint molecules~\cite{chem_similarity}.
        Resulting endpoint-midpoint Tanimoto distances for each compound class are shown in Figure \ref{fig:Interps}(b). \texttt{SALSA} generates interpolants that, on average, have a lower Tanimoto distance (therefore, are more similar) to their endpoints. This is indicative of improved semantic continuity in the \texttt{SALSA} space relative to the Naive space.

    \subsection{Property Awareness}

        \begin{figure}[t!]
            \centering
            \includegraphics[width=1.0\columnwidth]{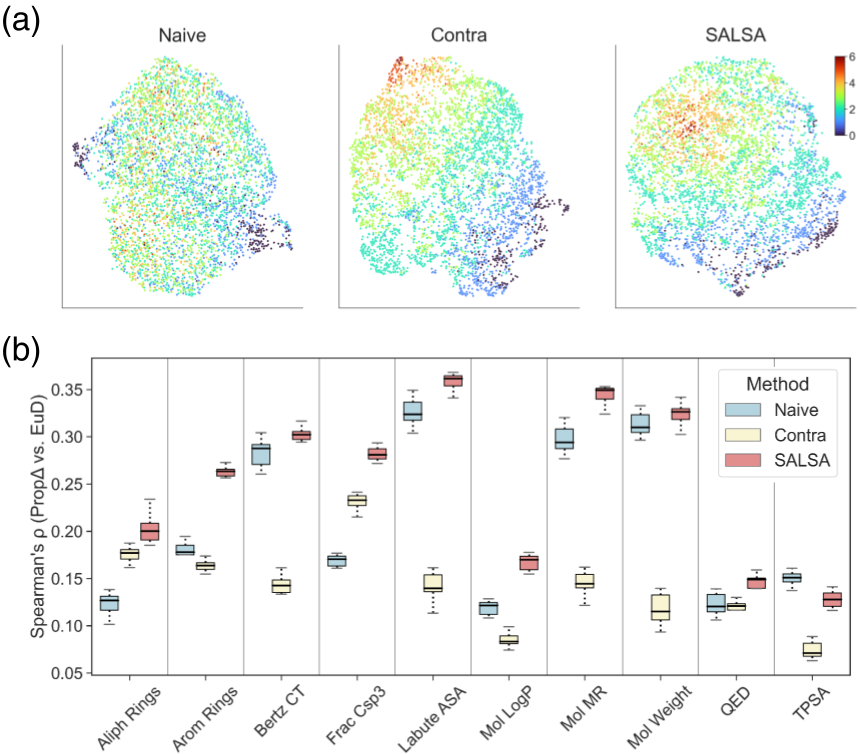}
            \caption{(a) UMAP reduction of 10,000 compounds, color-coded by Number of Aromatic Rings for Naive, Contra, and \texttt{SALSA} spaces. (b) Box plots of Prop$\Delta$-EuD Spearman's $\rho$ correlations across 10 physicochemical properties for Naive, Contra, and \texttt{SALSA} spaces.}
            \label{fig:PropUMAPs}
        \end{figure}
    
        In this work, we generally relate semantic awareness to structural awareness, but semantics may also be defined over chemical attributes, and furthermore, molecular structure often informs chemical attributes. Therefore, we find it pertinent to evaluate the extent to which latent representations capture information about physicochemical properties. To accomplish this we compute the correlation between property difference (Prop$\Delta$) and latent space Euclidean distance (EuD). This evaluation task is inspired from analogous tasks in NLP that correlate embedding similarity and human labels \cite{word_emb}.   

        We encode a sample of 2000 molecules into \texttt{SALSA}, Naive, and Contra space, and from those latent representations, compute the EuD between each pair for all three models. Following, we calculate 10 physicochemical properties (chosen for their relevance to drug discovery) for each molecule using RDKit \cite{feat_select_props, frac_csp3}. As an illustrative example, figure \ref{fig:PropUMAPs}(a) shows Uniform Manifold Approximation and Projection (UMAP) \cite{umap} reductions of a large set of 10,000 compounds in Naive, Contra, and \texttt{SALSA} space, color-coded by ``Number of Aromatic Rings''.
        For each property, we compute the property distance (Prop$\Delta$) between each molecular pair. Then, for each of the 10 properties across we compute the Spearman's rank correlation coefficient ($\rho$) between Prop$\Delta$ and EuD (for all three models). We perform this analysis on 10 random draws of 2000 molecules to obtain standard error; results are shown in Figure \ref{fig:PropUMAPs}(b). We find that \texttt{SALSA} achieves the highest correlation among models for nine out of 10 properties. 
        This is an intriguing finding as it is not obvious as to how \texttt{SALSA}'s framework enables better performance over either Naive or Contra. 
        One explanation could be that the contrastive loss works to sharply bring similar molecules close together, creating pockets of \textit{local} organization, while the reconstruction loss, again, enforces regularization such that the clusters disperse achieving more \textit{global} organization.

\section{Conclusion}
    In this work, we proposed \texttt{SALSA}, a framework for learning semantically meaningful latent representations. 
    Specifically, we sought to learn molecular representations informed by the structural similarities between molecules. 
    We trained a model with this intention and defined a direct evaluation metric, GED-EuD correlation, to show local structural awareness in latent space. Furthermore, we showed that \texttt{SALSA} produces more semantically reasonable interpolants, and that \texttt{SALSA} implicitly uncovers physicochemical properties, revealing wider context of latent organization. 
    Although we defined semantics in this work to be the \textit{structural} similarities between \textit{molecules}, the \texttt{SALSA} paradigm could be applied to any user-defined semantics based on \textit{x} similarity between \textit{y} data. In this way, \texttt{SALSA} could be potentially applied across a number of data types in various domains.

\section{Discussion}
     Beyond the scope of cheminformatics, we look to provide additional insight as to how \texttt{SALSA}'s methodological basis relates to a larger body of deep learning research. An interesting perspective from which to view our work is as a cousin to denoising adversarial autoencoders (DAAEs)~\cite{DBLP:journals/corr/MakhzaniSJG15}, particularly as applied to text or sequence data~\cite{shen2020educating}. The goal of the latter work, much like ours, is to coerce a sequence autoencoder to embed related sequences near one another. For the purposes of their DAAE, the natural data space metric is most closely related to a Levenshtein distance. While we seek to respect a different data space metric through \texttt{SALSA}, based on graph similarity, our goals are very much aligned with ~\citet{shen2020educating}. We opt for an objective function that, although distinct from that of ~\citet{shen2020educating}, we argue conceptually accomplishes a similar goal, nonetheless, to that of the DAAE objective. 
     % We leave to future work a more definitive theoretical analysis of our method.
     
    Our dual objective function for \texttt{SALSA} combines a \textit{reconstruction} loss and a \textit{contrastive} loss, which we claim acts similarly to the dual objective of the DAAE, combining a \textit{denoising} technique and an \textit{adversarial} loss. To support this claim, we refer to the work of~\cite{wang2020understanding}, wherein it was demonstrated that the contrastive loss, when restricted to latent vectors on the unit sphere and given the limit of infinite negative samples, simplifies into two components: an \textit{alignment} loss and a \textit{uniformity} loss. The alignment loss acts to align the latent representation of positive pairs, while the uniformity loss encourages the distribution of all latent vectors to be uniformly distributed on the unit sphere. Each of these losses has a conceptual counterpart in the DAAE, where the alignment loss acts similarly to the denoising objective and the uniformity loss acts like the adversarial component. In presenting this methodological comparison, we hope to provide a more general context for the techniques explored in \texttt{SALSA}, outside applications to molecular modeling.

% \pagebreak
\bibliography{aaai24}
%%%%%%%%%%%%%%%%%%%%%%%%%%%%%%%%%%%%%%%%%%%%%%%%%%%%%%%%%%%%%%%%%%%%%%
%%%%%%%%%%%%%%%%%%%%%%%%%%%%%%%%%%%%%%%%%%%%%%%%%%%%%%%%%%%%%%%%%%%%%%
% APPENDIX
%%%%%%%%%%%%%%%%%%%%%%%%%%%%%%%%%%%%%%%%%%%%%%%%%%%%%%%%%%%%%%%%%%%%%%
%%%%%%%%%%%%%%%%%%%%%%%%%%%%%%%%%%%%%%%%%%%%%%%%%%%%%%%%%%%%%%%%%%%%%%
\newpage
\appendix
\onecolumn

\section*{Funding}
This work was supported by the following grants to SMG from the National Institutes of Health - U24DK116204 (NIDDK), U01CA238475 (NCI), R01CA233811 (NCI), U01CA274298 (NCI). (https://www.nih.gov/). The funders had no role in study design, data collection and analysis, decision to publish, or preparation of the manuscript. During the course of this work, T.M.\ was supported by the National Institute of General Medical Sciences of the NIH under Award Number T32GM086330.

\end{document}